\documentclass[10pt,twocolumn,letterpaper]{article}

\usepackage{iccv}
\usepackage{times}
\usepackage{epsfig}
\usepackage{graphicx}
\usepackage{amsmath}
\usepackage{amssymb}
\usepackage{booktabs}
\usepackage{color, colortbl}
\usepackage{subcaption}
\usepackage{multirow}
\usepackage{pifont}
\usepackage{comment}
\newcommand{\cmark}{\ding{51}}%
\newcommand{\xmark}{\ding{55}}%
\usepackage{microtype}
\usepackage{threeparttable}
\usepackage[resetlabels,labeled]{multibib}
\usepackage[pagebackref=true,breaklinks=true,letterpaper=true,colorlinks,bookmarks=false]{hyperref}

\newcommand{\cmmnt}[1]{}

\usepackage{xspace}
\usepackage{cite}

\iccvfinalcopy 

\begin{document}


\title{Motion-Augmented Self-Training for Video Recognition at Smaller Scale}

\author{Kirill Gavrilyuk$^{1}$~~~~Mihir Jain$^{2}$~~~~Ilia Karmanov$^{2}$~~~~Cees G. M. Snoek$^{1}$\\[1mm]
\normalsize $^{1}$University of Amsterdam~~~~~$^{2}$Qualcomm AI Research\thanks{Qualcomm AI Research is an initiative of Qualcomm Technologies, Inc.}\\
{\tt\normalsize \{kgavrilyuk,cgmsnoek\}@uva.nl~~~~\{mijain,ikarmano\}@qti.qualcomm.com}
}

\maketitle

\begin{abstract}
The goal of this paper is to self-train a 3D convolutional neural network on an unlabeled video collection for deployment on small-scale video collections. As smaller video datasets benefit more from motion than appearance, we strive to train our network using optical flow, but avoid its computation during inference. We propose the first motion-augmented self-training regime, we call MotionFit. 
We start with supervised training of a motion model on a small, and labeled, video collection. With the motion model we generate pseudo-labels for a large unlabeled video collection, which enables us to transfer knowledge by learning to predict these pseudo-labels with an appearance model. 
Moreover, we introduce a multi-clip loss as a simple yet efficient way to improve the quality of the pseudo-labeling, even without additional auxiliary tasks. We also take into consideration the temporal granularity of videos during self-training of the appearance model, which was missed in previous works. 
As a result we obtain a strong motion-augmented representation model suited for video downstream tasks like action recognition and clip retrieval.
On small-scale video datasets, MotionFit outperforms alternatives for knowledge transfer by 5\%-8\%, video-only self-supervision by 1\%-7\% and semi-supervised learning by 9\%-18\% using the same amount of class labels. 
\end{abstract}

\section{Introduction} \label{sec:intro}

The goal of this paper is to self-train a 3D convolutional neural network on an unlabeled video collection, such that it can be effectively fine-tuned on small scale datasets. This is of interest for applications in small-sized companies, a household, or search and rescue robotics where large amounts of labeled video are often unavailable and the deployment in compute-efficient scenarios is preferred. The common self-training approach is to transfer knowledge from pre-trained appearance models by pseudo-label prediction, \eg~\cite{NorooziCVPR2018,  YanCVPR2020, rizve2021defense_iclr}. Yan \etal~\cite{YanCVPR2020}, for example, cluster a pre-learned appearance space before training a new network from scratch using cluster membership as pseudo-label. They transfer knowledge from 19 million weakly-labeled videos and use Kinetics~\cite{CarreiraCVPR2017} as their target with around 250k videos to fine-tune their model. 
Unlike them, we aim to self-train the model that can be effectively fine-tuned on small-scale datasets with around 10k videos. Such small video datasets benefit more from motion information than appearance \cite{SimonyanNIPS2014}, but the added flow computation affects the  efficiency. 
Other semi-supervised~\cite{rizve2021defense_iclr,jing2021videossl} and self-supervised~\cite{HanNIPS2020,HanArxiv2020} alternatives suitable for small-scale datasets either do not use motion~\cite{jing2021videossl,rizve2021defense_iclr} or use it at inference time also~\cite{HanNIPS2020,HanArxiv2020}.
So, we strive to train a convolutional neural network using optical flow, but avoid its computation during inference. We propose to transfer knowledge from the motion representation through self-training, to enable effective fine-tuning even on small-scale video collections.

\begin{figure}[t!]
\centering
 \includegraphics[width=0.92\linewidth]{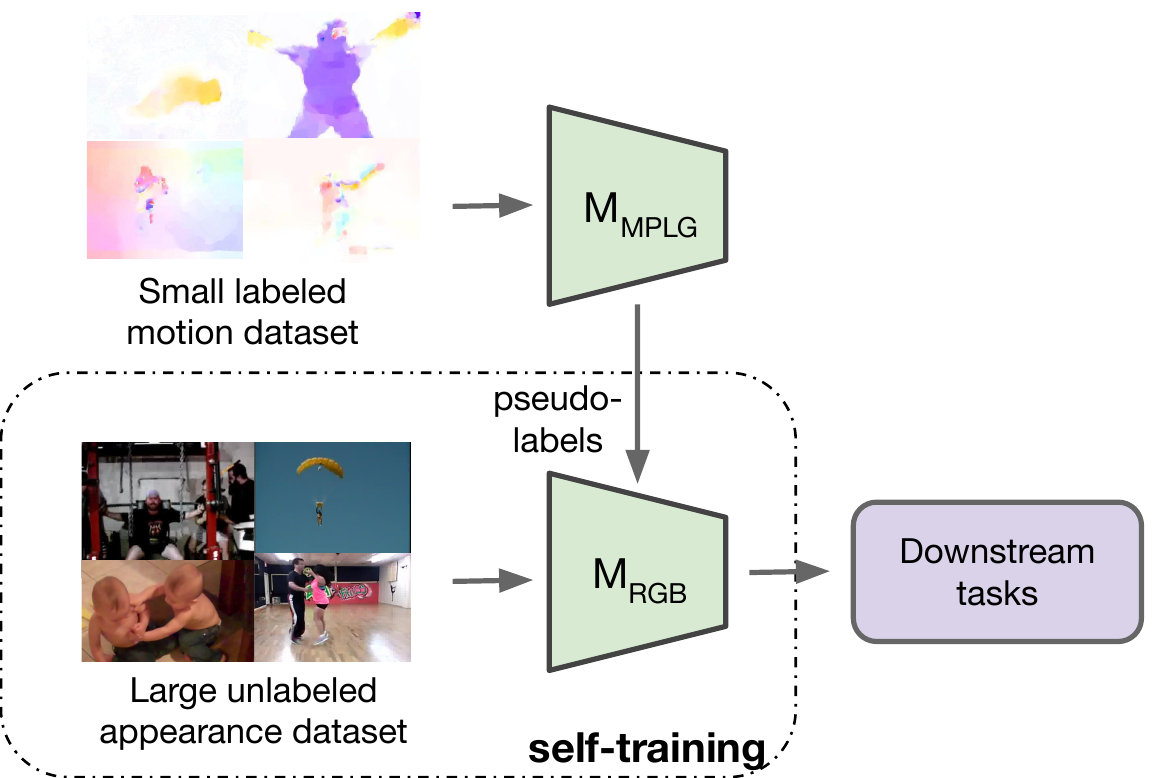}
\caption{\textbf{Motion-augmented self-training} utilizes pseudo-labels obtained by a motion model trained on a small labeled video dataset. We transfer knowledge from the motion model to the appearance model which is suitable for downstream video tasks without the need for optical flow computation.} 
\label{fig:spoiler}
\end{figure}

We are inspired by generalized distillation~\cite{LopezICLR2016}. During training it combines knowledge distillation~\cite{HintonArxiv2015} with privileged information~\cite{VapnikJMLR2015}. For example, to transfer knowledge from a pre-trained 2D convolutional neural network to a 3D convolutional neural network~\cite{DibaECCV2018, Girdhar2019ICCV} or from depth to an appearance stream~\cite{GarciaECCV2018, GarciaPAMI2019}. In particular, the works of Crasto \etal~\cite{CrastoCVPR2019} and Stroud \etal~\cite{StroudWACV2020} helped shape our idea. Both these works explore the transfer of motion knowledge to an appearance model using optical flow as privileged information, together with the large-scale labeled Kinetics~\cite{CarreiraCVPR2017} dataset. We also transfer from a motion to an appearance representation, but different from \cite{CrastoCVPR2019, StroudWACV2020} class labels on a large-scale dataset are \textit{unavailable} during transfer in our setting. Instead, we propose to obtain pseudo-labels by first training a motion model on a small-scale labeled dataset, like UCF101~\cite{SoomroArxiv2012} or HMDB51~\cite{KuehneICCV2011}. We perform unsupervised $K$-means clustering on the extracted motion features to obtain cluster assignments as pseudo-labels. Then we train an appearance model to predict these pseudo-labels via a self-training procedure on a larger source dataset, without using any additional class labels, see Figure~\ref{fig:spoiler}.

Our key contribution is a motion-augmented self-training procedure, we call \textit{MotionFit}. It extracts motion knowledge and transfers it to the appearance model, via self-training on a large-scale \textit{unlabeled} video dataset. By such motion transfer we avoid time-consuming optical flow computation during inference, similar in objective to motion knowledge distillation methods~\cite{CrastoCVPR2019, StroudWACV2020}, but without the need for labels during transfer.
Our second contribution is an empirical study to discover what form video pseudo-labels should take at smaller scale, starting from the training of the pseudo-label generator to temporally mapping pseudo-labels to videos. We train the pseudo-label generator on the motion representation of a small-scale and labeled video dataset by a multi-clip loss that makes our motion model less susceptible to the background motion irrelevant to the video label. During self-training with pseudo-labels we also study different levels of temporal video granularity by exploring several partitions of whole videos, which was not taken into account in related approaches, \eg,~\cite{ AlwasselArxiv2019,YanCVPR2020}.
Finally, we experimentally evaluate the importance of each component of our method and compare with state-of-the-art for action classification and clip retrieval on two datasets. 
For clip retrieval, we improve over the state-of-the-art on UCF101 and match it on HMDB51. For action classification, our self-trained representation performs considerably better than the alternative knowledge transfer (up to +8\%), self-supervised (up to +7\%) and semi-supervised (up to +18\%) methods.

\section{Related Work}\label{sec:related}

\textbf{Video self-training.}
The deep clustering of~\cite{CaronECCV2018} introduces an iterative approach by first assigning pseudo-labels using unsupervised $K$-means clustering, followed by predicting these assignments with a deep convolutional neural network. In~\cite{CaronICCV2019}, Caron \etal utilize a large non-curated dataset to train the deep clustering model. Asano \etal~\cite{AsanoICLR2020} suggest a principled learning formulation to overcome the problem of degenerate solutions of simultaneously learning and clustering features by maximizing the information between labels and input data indices. Zhan \etal~\cite{ZhanCVPR2020} propose an online deep clustering approach that performs clustering and network update simultaneously, rather than alternating. Both \cite{JiICCV2019} and \cite{CaronArxiv2020} align cluster assignments for pairs of different transformations of images.
A semi-supervised approach is presented in~\cite{rizve2021defense_iclr} where pseudo-labels are obtained directly from the predictions of a model trained on a labelled subset of the dataset. They propose a method for uncertainty-aware pseudo-label selection to iteratively select reliable unlabelled samples to be used with the labeled samples for training. Most similar to our work are Noroozi \etal~\cite{NorooziCVPR2018} and Yan \etal~\cite{YanCVPR2020}, who transfer knowledge from pre-trained models with self-training by pseudo-label prediction. Differently, we transfer knowledge from the motion representation to the appearance representation, while \cite{NorooziCVPR2018} and \cite{YanCVPR2020} exploit the same representation during pre-training and pseudo-label prediction. We also consider pseudo-labels on different levels of temporal video granularity, in an effort to better model the dynamic nature of video. 

\textbf{Video self-supervision.}
An alternative to self-training via pseudo-labeling is self-supervision.
Following the success of image-based self-supervision, early approaches for video self-supervision also explore similar pre-text tasks~\cite{JingArxiv2018, VondrickECCV2018, KimAAAI2019}. Other works pay more attention to the temporal video nature by exploiting chronological order~\cite{MisraECCV2016, FernandoCVPR2017, LeeICCV2017, XuCVPR2019}, pace~\cite{BenaimCVPR2020, YaoCVPR2020, ChoArxiv2020}, arrow of time~\cite{WeiCVPR2018} or video visual correspondence~\cite{KongNIPS2020, JabriNIPS2020}. The temporal structure is also beneficial for predicting future video states such as prediction of raw pixel representations of future frames~\cite{MathieuArxiv2016} or their feature representation~\cite{ElNoubyArXiv2019, HanICCVW2019, HanArxiv2020}. 

Most similar to our work are \cite{SayedGCPR2018, MahendranACCV2018, ZhanCVPR2019, HanArxiv2020, HanNIPS2020} who also explore optical flow. However, different from Sayed~\etal\cite{SayedGCPR2018} and Mahendran~\etal\cite{MahendranACCV2018}, who align features of appearance and motion representations, we align two representations via pseudo-label self-training. Zhan~\etal~\cite{ZhanCVPR2019} utilize sparse motion guidance to recover full-image motion from appearance while we predict just the same pseudo-label for both representations. Different from~\cite{HanArxiv2020} and~\cite{HanNIPS2020} who use flow in the two-stream fashion~\cite{SimonyanNIPS2014}, we still perform downstream tasks using only an appearance representation to be computationally efficient during inference. Differently from self-supervised methods we may make a second use of the small labeled target dataset to train the motion network for pseudo-label generation instead of only using it for fine-tuning the learned video representation model.

\begin{figure*}[t!]
    \centering
    \includegraphics[width=0.8\linewidth]{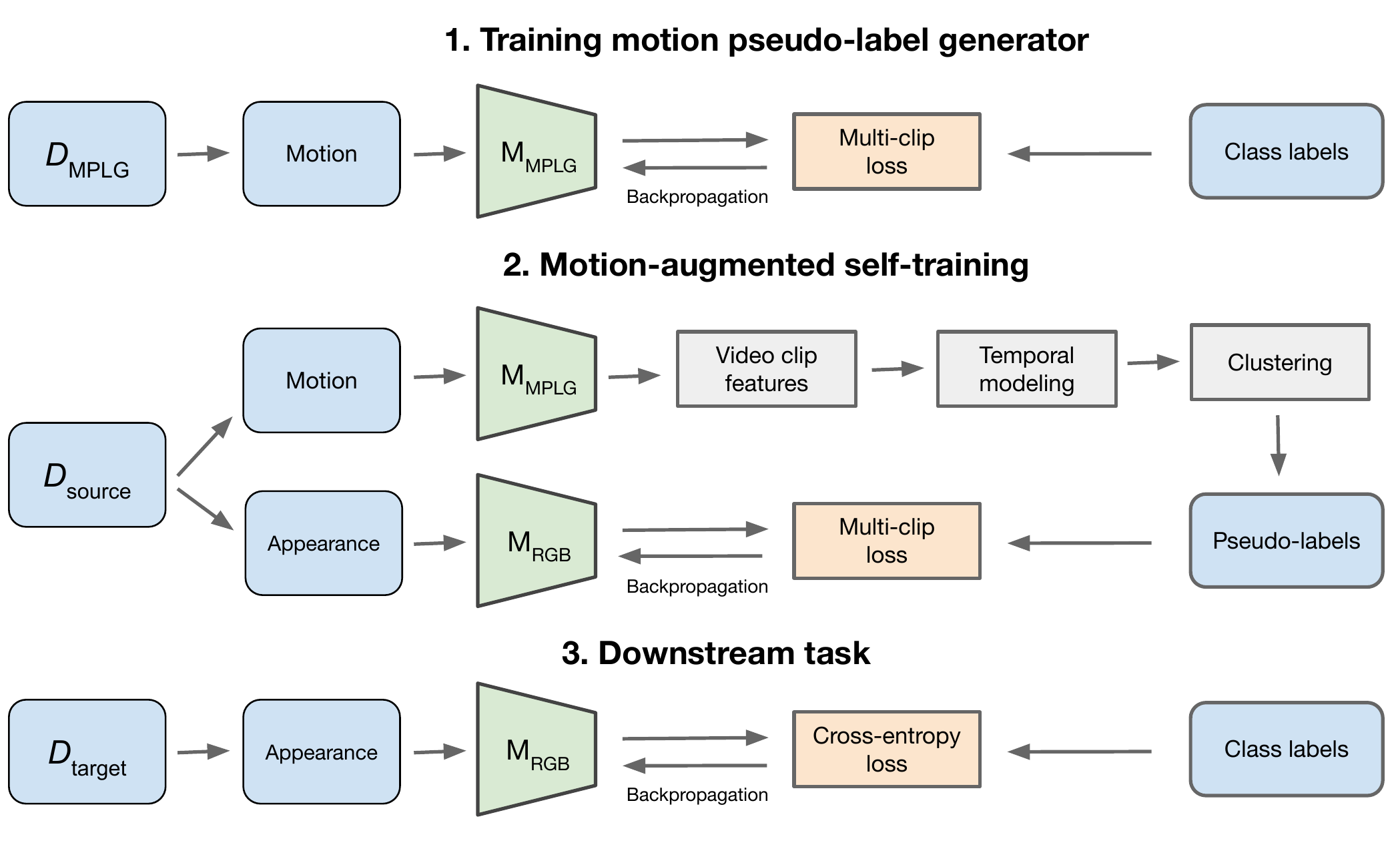}
    \caption{\textbf{MotionFit model.}
    Our approach consists of three main steps. In the first step we train a motion pseudo-label generator, $M_{\textit{MPLG}}$, on the motion representation of the dataset $D_{\textit{MPLG}}$ with a multi-clip loss. Next we use the $M_{\textit{MPLG}}$ to obtain pseudo-labels for the source dataset $D_{\textit{source}}$. We use these pseudo-labels for motion-augmented self-training of an appearance network $M_{\textit{RGB}}$ on $D_{\textit{source}}$, without any additional class labels. The learned video representation model is suitable for downstream tasks on $D_{\textit{target}}$ using only appearance as input. We do so by fine-tuning $M_{\textit{RGB}}$ on $D_{\textit{target}}$.
    }
    \label{fig:approach}
\end{figure*}

\textbf{Multimodal self-supervision.}
Many methods exploit the multimodal nature of videos, such as audio~\cite{OwensECCV2018, KorbarNIPS2018, AlwasselArxiv2019, PiergiovanniCVPR2020, MorgadoArXiv2020, PatrickArxiv2020, AsanoArxiv2020, AlayracNIPS2020} or corresponding text~\cite{LiArxiv2020} and speech~\cite{SunArxiv2019} by aligning multiple modalities between each other. Differently, we use an optical flow representation that can be derived from the raw RGB representation of the video. This representation helps to better model motion in the videos. Moreover, it does not introduce new information that is not contained in the original appearance representation, like audio and text.

\section{MotionFit Model}\label{sec:proposed}
Our goal is to learn a video representation suitable for downstream tasks, like video action recognition and video retrieval, on relatively small-scale target datasets $D_{\textit{target}}$ such as UCF101~\cite{SoomroArxiv2012} and HMDB51~\cite{KuehneICCV2011}. We first train a motion model, which we refer to as the motion pseudo-label generator $M_{\textit{MPLG}}$, also on a small-scale dataset $D_{\textit{MPLG}}$, which can be the same as $D_{target}$. Next, we use the $M_{\textit{MPLG}}$ network to extract motion features and to obtain pseudo-labels on a large-scale \textit{unlabeled} dataset $D_{\textit{source}}$. Then we switch to the RGB representation of $D_{\textit{source}}$ and self-train a new appearance network $M_{\textit{RGB}}$ to predict the pseudo-labels on $D_{\textit{source}}$ to obtain a motion-augmented video representation model suitable for downstream tasks on $D_{\textit{target}}$. The overall approach is illustrated in Figure~\ref{fig:approach} and detailed next. 

\subsection{Training the motion pseudo-label generator}
To train the motion pseudo-label generator on $D_{\textit{MPLG}}$ we can rely on the appearance representation by optimizing a cross-entropy loss to predict ground truth labels. However, current state-of-the-art networks are too large to be efficiently trained on small-scale datasets~\cite{CarreiraCVPR2017, XieECCV2018, TranCVPR2018, FeichtenhoferCVPR2019}. Instead, we pay more attention to the video nature and rely on a motion representation to train the $M_{\textit{MPLG}}$. The motion representation helps the convolutional neural network to concentrate on more important local motion changes, rather than the repetitive and abundant appearance representation. Hence, it has the advantage of learning from more generalizable information, while at the same time not heavily relying on context as the appearance representation does.

To further boost the strength of the $M_{\textit{MPLG}}$ representation we consider a longer temporal extent in each training sample, so it is more likely to contain the foreground part of the action of interest. This is important as unlike an appearance representation, motion cannot rely on background context for recognition. One way to cover longer temporal extents is to consider longer clips as input samples rather than the common practice of using 16-frame clips~\cite{XuCVPR2019, SunArxiv2019, KimAAAI2019, ZhuangCVPR2020}.
However, pooling over the temporal dimension can attenuate the foreground motion signal when a larger part is from the background. We propose an alternative strategy where each training sample consists of multiple clips, of standard 16-frames length, from the same video. The loss for each sample is obtained by averaging the cross-entropy losses over its clips.
The foreground clips are more likely to have a unimodal distribution across the final activations, while background ones  are likely to be relatively uniform. This means that averaging after the softmax (due to the nature of the exponential function) will produce higher logits and not attenuate the gradient information as much.
Thus, the proposed multi-clip loss preserves more temporal information and allows any clip from the foreground to contribute constructively to the back-propagating gradients.

\begin{align}
\mathcal{L}_{\textit{mc}}(y, \tilde{y}) = \frac{1}{B}\sum_{b=1}^{B}{\left( \frac{1}{R}\sum_{i=1}^{R}{\mathcal{L}_{\textit{class}}(y_{b,i}}, \tilde{y}_{b, i}) \right)},
\label{eq:proposed:loss_mc}
\end{align}
where $\mathcal{L}_{\textit{class}}$ is the cross-entropy loss for clip label prediction, $y_{b,i}$ is the ground truth clip label, $\tilde{y}_{b, i}$ is the model predicted clip label, $R$ is the number of sampled clips per video, and $B$ is batch size. In our experiments we show the benefits of this simple approach, which we call training with multi-clips.

Xu~\etal~\cite{XuCVPR2019} show that clip order prediction can be beneficial for self-supervised learning:
\begin{align}
\mathcal{L}_{\textit{co}}(p, \tilde{p})= \frac{1}{B}\sum_{b=1}^{B}{\mathcal{L}_{\textit{order}}(p_b, \tilde{p}_b)},
\label{eq:proposed:loss_co}
\end{align}
where $\mathcal{L}_{\textit{order}}$ is the cross-entropy loss for clip order prediction, $p_b$ is the correct order of the clips and $\tilde{p}_b$ is the model predicted clip order. In our multi-clip setting it is easy to combine it with our multi-clip loss $\mathcal{L}_{\textit{mc}}$ for more efficient training of the $M_{\textit{MPLG}}$. We do so by a weighted sum: 
\begin{align}
\mathcal{L}_{\textit{batch}} = \lambda\mathcal{L}_{\textit{mc}} + (1 - \lambda)\mathcal{L}_{\textit{co}},
\label{eq:proposed:loss}
\end{align}
where $\mathcal{L}_{\textit{batch}}$ is the batch loss, $\lambda\in[0;1]$ is a weighting parameter. With $\lambda {=} 0$ and using an appearance representation as input we have exactly the same self-supervised formulation as in~\cite{XuCVPR2019}. However, in our experiments we will show the benefits of using a motion representation and just relying on training with multi-clips.

\subsection{Motion-augmented self-training}
\label{sec:proposed:temporal_modeling}
Following~\cite{NorooziCVPR2018} and \cite{YanCVPR2020} we utilize $K$-means clustering with Euclidean distance to obtain pseudo-labels for the source dataset $D_{\textit{source}}$. Rather than relying on appearance features as~\cite{NorooziCVPR2018} and~\cite{YanCVPR2020} do, we do so on the motion features extracted by our $M_{\textit{MPLG}}$. The cluster centers computed by $K$-means are considered as pseudo-labels for the previously unseen videos from the $D_{\textit{source}}$ dataset. We train another network $M_{\textit{RGB}}$, but with an appearance representation as input, to predict the pseudo-labels of $D_{\textit{source}}$. By doing so, we transfer motion knowledge from the $M_{\textit{MPLG}}$ to the appearance network $M_{\textit{RGB}}$. Noroozi~\etal~\cite{NorooziCVPR2018} and Yan~\etal~\cite{YanCVPR2020} show that this self-training procedure is the most efficient way to do so, compared to other model distillation and transfer techniques. However, where they rely on RGB appearance only, we consider different video representations. We train the $M_{\textit{MPLG}}$ on motion and network $M_{\textit{RGB}}$ on appearance. The motion representation allows us to train the $M_{\textit{MPLG}}$ using only a small-scale dataset, while the appearance representation of $M_{\textit{RGB}}$ is computationally efficient during inference time. Note that $M_{\textit{MPLG}}$ and $M_{\textit{RGB}}$ can also have a different architecture, allowing us to train network $M_{\textit{RGB}}$ with different backbones using the same set of pseudo-labels obtained by $M_{\textit{MPLG}}$. In the experiments we also compare with other knowledge transfers including transfer from pre-trained 2D CNNs~\cite{DibaECCV2018, Girdhar2019ICCV} and  motion~\cite{CrastoCVPR2019, StroudWACV2020}, in either case showing the benefits of our approach.

\textbf{Temporal modeling for self-training.} 
To further enrich the representation learned by $M_{\textit{RGB}}$, we use a similar loss as Equation~\ref{eq:proposed:loss} to train $M_{\textit{RGB}}$ on $D_{\textit{source}}$. Different from~\cite{AlwasselArxiv2019, YanCVPR2020} we also consider assigning pseudo-labels to subparts of the video, taking into account the most suitable temporal scale for the pseudo-labels. To do so, we first extract features using a trained $M_{\textit{MPLG}}$ for each video clip that we densely sample from all videos $V {=} \{V_{i}\}_{i{=}1}^{N_{\textit{source}}}$ of the dataset $D_{\textit{source}}$: $F_{i} {=} \{f_{i,t}\}_{t{=}1}^{T_{i}}$, where $T_i$ is number of clips in $V_i$; $N_{\textit{source}}$ is the number of videos in $D_{\textit{source}}$. Video clip is a sequence of adjacent frames in the video; 16, 32 or 64 frames in our experiments. After this we consider three levels of temporal granularity: clip level, segment level and video level. These levels differ in the way they aggregate features using $F_{i}$ to obtain new $H_i{=}\{h_{i, s}\}_{s{=}1}^{S}$ representation for each video $V_i$. 

For clip level we consider $H_{i} {=} F_{i}$. We define a segment as a sequence of adjacent video clips and consider two methods to obtain them. For the first, we utilize the same approach as in~\cite{JainCVPR2020}. For a given video, we find segment boundaries $B_i$ by looking for time steps where features of adjacent clips change abruptly compared to the previous time step:
\begin{align}
B_i=\{t: || f_{i, t} - f_{i, t-1}||_1 > \tau\}
\label{eq:proposed:ab}
\end{align}
where $\tau$ is set to the $p$-th percentile, so the number of segments in a video $V_i$ is directly proportional to its length $T_i$. For the second, we consider the approach where the video is equally divided into three segments similar to~\cite{WangECCV2016}. Following~\cite{JainCVPR2020}, to obtain the feature representation for each segment we average clip features within its boundaries. For video level we consider the whole video as one segment and represent it by also averaging features of all clips in the video similar to~\cite{YanCVPR2020}. Independent of the level of temporal granularity the other parts of our method are kept unchanged. 

\section{Experiments}\label{sec:experiments}

\subsection{Datasets}\label{sec:experiments:datasets}
We validate our approach using the following three datasets\footnote{Datasets used in this paper were downloaded and experimented on by primary author}. 

\textbf{UCF101 and HMDB51}. As instances of our $D_{\textit{MPLG}}$ and $D_{\textit{target}}$ datasets we rely on UCF101~\cite{SoomroArxiv2012} and HMDB51~\cite{KuehneICCV2011}, two well-known datasets for video action recognition. However, in many works~\cite{CarreiraCVPR2017, XieECCV2018, TranCVPR2018, FeichtenhoferCVPR2019} it is shown that these two datasets are too small for an efficient training of modern 3D CNNs on the appearance representation. Therefore these datasets are good choice for our experiments. UCF101 contains 13k videos with 101 human actions. There are 3 splits available with around 9k training and 4k testing videos per split. HMDB51 contains 7k videos with 51 human classes from movies. It also has 3 splits with around 5.5k training and 1.5k testing videos per split. We consider these datasets for both downstream tasks of video action recognition and video clip retrieval.

\textbf{Kinetics-400}. As instance of $D_{\textit{source}}$, we rely on Kinetics-400~\cite{CarreiraCVPR2017}. It contains 400 human action classes with 10-second clips. There are around 246k training and 50k validation videos. Kinetics-400 is considered as a standard choice for pre-training of modern 3D CNNs~\cite{CarreiraCVPR2017, XieECCV2018, TranCVPR2018, FeichtenhoferCVPR2019}. Similar to self-supervised methods we use this dataset without human annotation to pre-train 3D CNNs. We perform ablation experiments on its validation split (Kinetics-val) and make comparison with knowledge transfer and self-supervised methods using its train split (Kinetics-train). 

\subsection{Implementation details} \label{sec:experiments:implementation}
\textbf{Motion pseudo-label generator}. For our motion representation we utilize TV-L1~\cite{ZachDAGM2007}, which is widely used for video action recognition. Following the recent literature~\cite{AsanoArxiv2020,CrastoCVPR2019,PatrickArxiv2020}, we choose the R(2+1)D-18~\cite{TranCVPR2018} as the backbone for our $M_{\textit{MPLG}}$. 

\textbf{Self-training}. For appearance model $M_{\textit{RGB}}$, we consider again the R(2+1)D-18 backbone as well as the S3D-G~\cite{XieECCV2018}.

\textbf{Training details}. We randomly split around 10\% videos from the training set to do validation both during training $M_{\textit{MPLG}}$ and self-training. The input video clips are first resized to $128 {\times} 171$, then randomly cropped to $112 {\times} 112$ during training. During validation or testing, the clip is cropped in the center. For S3D-G, we use longer clips with 64 frames and higher resolution of $224 {\times} 224$. Recent findings \cite{masters2018revisiting} show that small mini-batch sizes provide more up-to-date gradient calculations and yield more stable and reliable training. Thus we use 8 multi-clips per batch while training on smaller datasets. For self-training on Kinetics-400 we use a batch size of 16 for R(2+1)D and 32 for S3D-G. We train our model with an initial learning rate of 0.001 and decay the learning rate by a factor of 10 at steps of 40, 60, 80 epochs. This regime is followed during $M_{\textit{MPLG}}$ training and fine-tuning (only conv4, conv5 and fc layers) for each dataset. For self-training, we decay the learning rate at 20 and 40 epochs.
The $M_{\textit{MPLG}}$ training and fine-tuning is done for up to 120 epochs, while self-training on the Kinetics-400 train set is done for 45 epochs.
Our models are implemented with PyTorch and optimized with vanilla synchronous SGD algorithm with momentum of 0.9 and weight decay of 0.0005. 

\begin{table}[t!]
\centering
\resizebox{0.85\linewidth}{!}{
\begin{tabular}{lccccc}
\toprule
 & \multicolumn{2}{c}{\textbf{Clip length}} & \multicolumn{3}{c}{\textbf{Multi-clips ($R$)}}\\
\cmidrule(lr){2-3} \cmidrule(lr){4-6}
\bfseries Representation & 32 &  64 &  1 &  2 &  3 \\
\midrule
Appearance & 59.4 & 60.3 & 58.9 & 57.0 & 59.1\\
Motion & 80.8 & 81.1 & 78.2 & 82.2 & 82.6 \\
\bottomrule
\end{tabular}
}
\caption{\textbf{Benefit of $M_{\textit{MPLG}}$ motion representation.} Comparison between appearance and motion representation for varying clip length (in frames) and multi-clips per video on UCF101 split 1. When using multiple clips for training we consider 16-frames clips as input. Using multi-clip training is even better than using larger input temporal extent which allows us to efficiently train $M_{\textit{MPLG}}$ without additional computational and memory cost. The motion representation is at least 20\% better than appearance, for all settings.}
\label{table:experiments:ablation_representation}
\end{table}

\subsection{Ablation of motion pseudo label generator}
\label{sec:experiments:ablation_MPLG}
\textbf{Benefit of $M_{\textit{MPLG}}$ motion representation}. We first perform an ablation to demonstrate the benefit of training the pseudo-label generator on the optical flow representation. We report action recognition results on UCF101 split 1 for this experiment. First, we train the $M_{\textit{MPLG}}$ with the loss function in Eq.~\ref{eq:proposed:loss}, setting $\lambda{=}1$ to show only the importance of using multiple clips from the same video in the batch. In Table~\ref{table:experiments:ablation_representation} we ablate appearance and motion representations for a varying number of clips $R$ in Eq.~\ref{eq:proposed:loss}. We also compare with standard single clip training, but with longer temporal length of the input clip. Independent of the clip length and number, the motion representation is at least 20\% better than appearance, even when using a small temporal scale. Our multi-clip training is helpful for motion representation learning giving around 4\% boost over standard single-clip training. It is also more than 1\% better than using clips with larger temporal scale. While for appearance, which relies heavily on context, there is not much improvement when using longer temporal scale or multiple clips for training.

\textbf{Choice of $M_{\textit{MPLG}}$ parameter $\lambda$}. Next we perform experiments varying parameter $\lambda$ in Eq.~\ref{eq:proposed:loss}. While for the appearance representation the improvement matches the results of~\cite{XuCVPR2019} (73.7) the motion representation does not benefit from explicit clip order prediction. With $\lambda{=}0$ we achieve only 69.9 accuracy, which is even lower than for the appearance representation. For values of $\lambda {<} 1$ we  did not see any gain in action recognition accuracy compared to the case of $\lambda {=} 1$. This supports that our multi-clip training procedure by itself provides enough temporal information when trained with a motion representation. Therefore, we use a motion representation for our $M_{\textit{MPLG}}$ multi-clip training with $R{=}3$ and $\lambda{=}1$ in the rest of the paper.   

\begin{table}[t!]
\centering
\resizebox{0.9\linewidth}{!}{
\begin{tabular}{cccc}
\toprule
\multicolumn{4}{c}{\bfseries Temporal granularity} \\
\cmidrule(lr){1-4} 
Video~\cite{AlwasselArxiv2019, YanCVPR2020} & Segment~\cite{JainCVPR2020} & Segment~\cite{WangECCV2016} & Clip \\
\midrule
76.5 & 79.0 &  77.3 & 80.3 \\
\bottomrule
\end{tabular}
}
\caption{\textbf{Choice of temporal granularity} for motion-augmented self-training of model $M_{\textit{RGB}}$ on Kinetics-val. Comparison is performed on the downstream task of video action recognition on UCF101 split 1. Clip level gives ~4\% boost over video level and is slightly better than segments.}
\label{table:experiments:ablation_temporal}
\end{table}

\textbf{Importance of $M_{\textit{MPLG}}$ for self-training}. 
First, we use the above $M_{\textit{MPLG}}$ ($R{=}3$, $\lambda{=}1$) to generate pseudo-labels (128 clusters) for self-training on Kinetics-400 train set, and obtain 85.2\% accuracy on UCF101.
Then, we perform the same experiment but with $M_{\textit{MPLG}}$ trained on appearance instead.
By self-training on these appearance pseudo-labels, we obtain only 74.5\%, which confirms the better quality of our motion pseudo-labels. 
Note that the setting with appearance pseudo-labels is similar to~\cite{NorooziCVPR2018} and \cite{YanCVPR2020}, which are developed for the image domain and very large-scale datasets, respectively. Unlike them, we effectively train our model for small-scale video datasets using motion, but only relying on appearance during inference.

\subsection{Ablation of motion-augmented self-training} \label{sec:experiments:self_training}
Next we ablate the self-training choices for the network $M_{\textit{RGB}}$, where we use the Kinetics-400 validation set as our source dataset $D_{\textit{source}}$. We train our $M_{\textit{MPLG}}$ on UCF-101 split 1 with the motion representation, and use it to extract features for densely sampled clips from each video in $D_{\textit{source}}$. We compare the choices of the self-trained appearance model $M_{\textit{RGB}}$ on the downstream task of video action recognition on UCF-101 split 1. Note that during self-training the model $M_{\textit{RGB}}$ does not see any videos from UCF-101 neither any provided class labels of the Kinetics-400 dataset.

\textbf{Choice of temporal granularity}. We first analyze the importance of considering temporal scale for generating pseudo-labels, as discussed in Section~\ref{sec:proposed:temporal_modeling}. We show the results in Table~\ref{table:experiments:ablation_temporal} for three possible levels: video, segment and clip. Interestingly, a more semantic partition of videos as suggested in~\cite{JainCVPR2020} and~\cite{WangECCV2016} does not improve the representation of $M_{\textit{RGB}}$ compared to just a clip-level granularity. However, both segment and clip levels outperform video level~\cite{AlwasselArxiv2019, YanCVPR2020} up to 2.5\%. We also vary parameter $\lambda$ in Eq.~\ref{eq:proposed:loss} during self-training of $M_{\textit{RGB}}$. For none of the temporal levels we have seen any considerable improvement over using $\lambda {=} 1$. To further investigate the role of clip order we add the loss for pseudo-label prediction of the next and previous clips for each sampled clip of the video. This gives us an additional 0.4\% improvement. These results again support that a self-training procedure with just a multi-clip loss helps successfully transfer motion knowledge to the appearance model, even without any additional modeling of clip order.  

\begin{table}[t!]
\centering
\resizebox{0.8\linewidth}{!}{
\begin{tabular}{lcccc}
\toprule
& \multicolumn{4}{c}{\textbf{Number of clusters}}\\
\cmidrule(lr){2-5}
& 128 & 500 & 1000 & 1600\\
\midrule
Kinetics-val    & 79.0 & 79.0 & 79.7 & 74.2\\
Kinetics-train  & 85.2 & 85.7 & 86.5 & 85.6\\
\bottomrule
\end{tabular}
}
\caption{\textbf{Choice of number of clusters} for self-training of model $M_{\textit{RGB}}$ on the Kinetics-val and Kinetics-train. Comparison is performed on the downstream task of video action recognition on UCF101 split 1. A larger number of clusters benefits self-training up to 1000 clusters.}
\label{table:experiments:ablation_clusters}
\end{table}

\textbf{Choice of number of clusters}. Next we compare the influence of the number of clusters used in $K$-means to obtain pseudo-labels for the source dataset $D_{\textit{source}}$ in Table~\ref{table:experiments:ablation_clusters}. When using Kinetics-400 validation for self-training of $M_{\textit{RGB}}$ the impact of using more clusters is minimal. Even on the larger Kinetics-400 train set
there is only small improvement of up to 1.3\%, indicating our motion-augmented self-training is not sensitive to the choice of number of clusters. We choose $K{=}1000$ for all the following experiments.

\textbf{Effect of common classes}. UCF101 and Kinetics-400 have 55 classes in common. To evaluate their impact, we exclude them from the latter and re-train MotionFit ($K{=}1000$) with the remaining $\sim$170k videos. We obtain 86.3\% on split1 of UCF101 \textit{vs.} 86.5\% with all the classes, which shows the common classes have hardly an effect on accuracy.

\begin{table}[t!]
\centering
\resizebox{0.8\linewidth}{!}{
\begin{tabular}{lcc}
\toprule
& \multicolumn{2}{c}{{\textbf{$D_{\textit{target}}$}}}\\
\cmidrule(lr){2-3}
\textbf{Clip-length} & UCF101 & HMDB51\\
\midrule
16 frames & 87.4 & 56.4 \\ 
32 frames & 88.9 & 61.4 \\
\midrule
Random initialized~\cite{XuCVPR2019} & 58.9 & 22.0 \\
\bottomrule
\end{tabular}
}
\caption{\textbf{Impact on target dataset} for varying clip-length, where $D_{\textit{MPLG}}$ is UCF101 and $D_{\textit{source}}$ is Kinetics. Average accuracy over three splits is reported.}
\label{table:experiments:ablation_target}
\end{table}

\begin{table*}[t!]
\centering
\resizebox{0.78\linewidth}{!}{
\begin{threeparttable}
\begin{tabular}{lcccc|cc}
\toprule
& \bfseries Backbone & \bfseries Frames & \bfseries Resolution & \bfseries Additional labels & \bfseries UCF101 & \bfseries HMDB51\\
\midrule
Random initialisation~\cite{XuCVPR2019} & R(2+1)D-18 & 16 & 112 & -- & 58.9 & 22.0 \\
MERS~\cite{CrastoCVPR2019}$^{\dagger}$ & R(2+1)D-18 & 16 & 112 & -- & 78.3 & 42.1 \\
MARS~\cite{CrastoCVPR2019}$^{\dagger}$ & R(2+1)D-18 & 16 & 112 & -- & 82.2 & 48.7 \\
STC~\cite{DibaECCV2018} & STC-ResNext & 16 & 112 & ImageNet & 84.7 & - \\
DistInit~\cite{Girdhar2019ICCV} & R(2+1)D-18 & 32 & 112 & ImageNet & 85.7 & 54.9 \\
Supervised~\cite{PatrickArxiv2020} & R(2+1)D-18 & 16 & 112 & Kinetics-400 & 95.0 & 70.4 \\
\midrule
\textbf{MotionFit} (\textit{ours}) & R(2+1)D-18 & 16 & 112 & -- & 87.4 & 56.4 \\
\bottomrule
\end{tabular}
\footnotesize{$^{\dagger}$MERS and MARS results are based on our implementation of~\cite{CrastoCVPR2019}}. 
  \end{threeparttable}
  }
\caption{\textbf{Comparison with knowledge transfer} methods on video action recogntion. We report top-1 accuracy of fine-tuned models averaged over all 3 splits of UCF101 and HMDB51. Our approach outperforms MERS and MARS for transferring motion knowledge to the appearance stream. MotionFit is also better than methods that also rely on ImageNet class labels.}
\label{table:experiments:init_comparison}
\end{table*}

\begin{table*}[t!]
\centering
\resizebox{0.78\linewidth}{!}{
\begin{tabular}{lcccc|cc}
\toprule
& \bfseries Backbone & \bfseries Frames & \bfseries Resolution & \bfseries Modality & \bfseries UCF101 & \bfseries HMDB51\\
\bottomrule
Sun~\etal~\cite{SunArxiv2019} & S3D & 16 & 112 & V + T & 79.5 & 44.6 \\
Asano~\etal~\cite{AsanoArxiv2020} & R(2+1)D-18 & 30 & 112 & V + A & 83.1 & 47.1 \\
Alwassel~\etal~\cite{AlwasselArxiv2019} & R(2+1)D-18 & 32 & 224 & V + A & 86.8 & 52.6 \\
Xiao~\etal~\cite{XiaoArxiv2020} & SlowFast & 64 & 224 & V + A & 87.0 & 54.6 \\
Morgado~\etal~\cite{MorgadoArXiv2020} & R(2+1)D-18 & 32 & 224 & V + A & 87.5 & 60.8 \\
Patrick~\etal~\cite{PatrickArxiv2020} & R(2+1)D-18 & 32 & 224 & V + A & 89.3 & 60.0 \\
\cmidrule{1-7}
Kim~\etal~\cite{KimAAAI2019} & R3D-18 & 16 & 112 & V & 65.8 & 33.7 \\
Kong~\etal~\cite{KongNIPS2020} & R3D-18 & 8 & 112 & V & 69.4 & 37.8 \\
Han~\etal~\cite{HanICCVW2019} & R-2D3D-34 & 25 & 224 & V & 75.7 & 35.7 \\
Jing~\etal~\cite{JingArxiv2018} & R3D-18 & 64 & 112 & V & 76.6 & 47.0 \\
Zhuang~\etal~\cite{ZhuangCVPR2020} & SlowFast & 16 & 112 & V & 77.0 & 46.5 \\
Han~\etal~\cite{HanArxiv2020} & R-2D3D-18 & 25 & 224 & V & 78.1 & 41.2 \\
Benaim~\etal~\cite{BenaimCVPR2020} & S3D-G & 64 & 224 & V & 81.1 & 48.8 \\
Han~\etal~\cite{HanNIPS2020} & S3D & 32 & 128 & V & 87.9 &  54.6 \\
\cmidrule{1-7}
\textbf{MotionFit} (\textit{ours}) & R(2+1)D-18 & 32 & 112 & V & 88.9 & 61.4 \\
\textbf{MotionFit} (\textit{ours}) & S3D-G & 64 & 224 & V & 90.1 & 50.6 \\
\bottomrule
\end{tabular}
}
\caption{\textbf{Comparison with self-supervised methods on video action recognition.} We report top-1 accuracy of fine-tuned models averaged over all 3 splits of UCF101 and HMDB51. For fair comparison we list only the results obtained using the Kinetics-400 dataset. Our approach is the best when only considering the visual modality (V). On UCF101 we are even on par with methods that use an additional audio modality (V+A) during training. }
\label{table:experiments:action_state_of_the_art}
\end{table*}

\textbf{Impact on target dataset.} 
Our self-training approach trains an appearance model with motion information to be effectively fine-tuned on small video datasets.
First we evaluate our method for the case when $D_{\textit{MPLG}}$ and $D_{\textit{target}}$ are the same \ie UCF101, and $D_{\textit{source}}$ is always Kinetics. We fine-tune and evaluate on each of the three splits of UCF101 and  report the results in the first row of Table~\ref{table:experiments:ablation_target}. For self-training we use 16-frame clips, however, increasing the clip-length to 32 for fine-tuning leads to improved performance, with little  additional computation. 
Then, we change the $D_{\textit{target}}$ to HMDB51 but keep the same self-trained model. Again, there is considerable accuracy gain over random initialization~\cite{BenaimCVPR2020}, and longer clips help, but the main conclusion is that our approach can be easily applied to other small datasets without requiring the whole pipeline to be redone (flow extraction and then self-training on Kinetics). A considerable practical advantage. Next, we compare with the state-of-the-art.

\subsection{Comparisons with state-of-the-art} \label{sec:experiments:comparison_knowledge_action}

\textbf{Knowledge transfer.}
In Table~\ref{table:experiments:init_comparison} we compare with knowledge transfer methods. As knowledge transfer methods we consider two main families: transfer knowledge from pre-trained 2D CNN and knowledge transfer from the motion to the appearance stream, the same as we do. For the former we choose STC~\cite{DibaECCV2018} and DistInit~\cite{Girdhar2019ICCV}, for the latter we choose MERS and MARS from~\cite{CrastoCVPR2019}.
We train MERS by matching features of the student appearance network with the teacher motion network on Kinetics-400. Then we further fine-tune the student model on the target dataset as we did for MotionFit. MARS additionally combines the feature matching with a cross-entropy loss and needs class-labels, so it is trained directly on the target dataset. For the teacher network we use the same $M_{\textit{MPLG}}$ used to obtain pseudo-labels to train MotionFit. We have a good improvement ($>$5\% on both datasets for R(2+1)D-18) over motion transfer methods that are based on feature matching. We are even better than knowledge transfer approaches from pre-trained 2D CNN, despite them also using ImageNet class labels.

\begin{table*}[t!]
\centering
\resizebox{0.75\linewidth}{!}{
\begin{tabular}{lccccccccc}
\toprule
& & & \multicolumn{3}{p{3.0cm}}{\centering\bfseries UCF101} & \multicolumn{3}{p{3.0cm}}{\centering\bfseries HMDB51} \\
\cmidrule(lr){4-6} \cmidrule(lr){7-9}
& {\bfseries Backbone} & {\bfseries Modality} & R@1 & R@5 & R@20 & R@1 & R@5 & R@20 \\
\midrule
Benaim~\etal~\cite{BenaimCVPR2020} & S3D-G & V & 13.0 & 28.1 & 49.5 & - & - & - \\
Kong~\etal~\cite{KongNIPS2020} & R3D-18 & V & 22.0 & 39.1 & 56.3 & - & - & - \\
Asano~\etal~\cite{AsanoArxiv2020} & R(2+1)D-18 & V + A & 52.0 & 68.6 & 84.5 & 24.8 & 47.6 & 75.5 \\
Patrick~\etal~\cite{PatrickArxiv2020} & R(2+1)D-18 & V + A & 57.4 & 73.4 & 88.1 & 25.4 & 51.4 & 75.0 \\
\midrule
\textbf{MotionFit} (\textit{ours}) & S3D-G & V & 31.6 & 51.7 & 70.3 & - & - & - \\
\textbf{MotionFit} (\textit{ours}) & R(2+1)D-18 & V & 61.6 & 75.6 & 85.5 & 29.4 & 46.5 & 66.7 \\
\bottomrule
\end{tabular}
}
\caption{\textbf{Comparison with self-supervised methods on video clip retrieval.} We report recall values R@$n$ for $n$ = 1, 5, 20 on UCF101 and HMDB51 split 1. For fair comparison we list only the results obtained using Kinetics-400. Our approach is best when only considering the visual modality and on par with methods that use an additional audio modality during training.}
\label{table:experiments:retrieval_state_of_the_art}
\end{table*}

\textbf{Self-supervised action recognition.}
Next, we compare our approach with state-of-the-art self-supervised methods as we also do not use ground truth labels during self-training on the Kinetics-400 dataset.
We first compare on video action recognition on UCF101 and HMDB51 in Table~\ref{table:experiments:action_state_of_the_art}. Note that all reported methods fine-tune their models on these datasets and hence use the same amount of class labels as we do. These methods vary vastly for the choice of backbone networks, number of input frames and input resolutions, making fair comparison a challenging task by itself. However, thanks to the simplicity of our method, we can easily train different backbone appearance networks $M_{\textit{RGB}}$ to predict the pseudo-labels generated by the same $M_{\textit{MPLG}}$.
We report for two backbone networks and outperform most other methods that use only the video modality with good margins on both the datasets. Using the same S3D-G backbone as Benaim~\etal~\cite{BenaimCVPR2020}, we obtain an increase of 9.0\% on UCF101 and 1.8\% on HMDB51. We are also slightly better than Han~\etal~\cite{HanNIPS2020} on UCF101 who use an S3D backbone.
We even perform better than some multi-modal methods that additionally utilize text~\cite{SunArxiv2019} or audio~\cite{AsanoArxiv2020}. For similar resolution and number of input frames, we are almost on par with most multi-modal methods. 
We conclude that our approach makes a better use of the class labels of small  datasets due to motion-augmented self-training.

\textbf{Self-supervised clip retrieval}. Next, we compare with the state-of-the-art self-supervised methods on video clip retrieval in Table~\ref{table:experiments:retrieval_state_of_the_art}. We follow the setting of Xu \etal~\cite{XuCVPR2019} and use split 1 of UCF101 and HMDB51 for comparison. From each video we sample 10 clips and clips extracted from the testing set are used to query the clips from the training set. We use max-pooled features after the last residual block as a clip feature representation. 
For a query clip, $n$ nearest training clips are retrieved, and if any of them has the same class-label as the query 
the retrieval is deemed correct.
We report recall results at different values of $n$. Our method outperforms the method of Benaim \etal~\cite{BenaimCVPR2020} by a large margin for both the backbone networks. We are on par with the methods of Asano \etal~\cite{AsanoArxiv2020} and Patrick \etal~\cite{PatrickArxiv2020}, which both leverage an additional audio modality. We conclude that our motion-augmented self-trained model $M_{\textit{RGB}}$ learns distinctive action motions despite being trained only on the appearance representation, see also qualitative retrieval results in the supplemental.  

\textbf{Semi-supervised action recognition.}
Finally, in Table~\ref{table:experiments:semis}, we compare with semi-supervised methods for video recognition on a small-scale labelled dataset. Following Jing~\etal~\cite{jing2021videossl} and Rizve~\etal~\cite{rizve2021defense_iclr}, we experiment on split 1 of UCF101 and use 3D ResNet-18 as backbone. 
For their semi-supervised learning, the competing methods at random select a fraction (20\% or 50\%) of the train set as labelled subset and use the rest of the videos without labels. The model is trained on the unlabeled as well as the labeled subsets.
For fair comparison, we set the labelled fraction as $D_{\textit{MPLG}}$ (same as $D_{\textit{target}}$), and the full train set without labels as $D_{\textit{source}}$. 
Despite Jing~\etal using extra labels to pre-train a 2D CNN, we outperform them by 9\% and 5.7\% for 20\% and 50\% labelled data, respectively. The gain over Rizve~\etal is even more, around 18.3\% and 8.8\% for the two labelled subsets. As the number of labelled videos decrease, we observe further advantage of MotionFit. More experimental analysis is reported in the supplementary material.

\begin{table}[t!]
\centering
\resizebox{0.85\linewidth}{!}{
\begin{threeparttable}
\begin{tabular}{lcc}
\toprule
& \multicolumn{2}{c}{\textbf{UCF101, split 1}}\\
\cmidrule(lr){2-3}
 & 20\% labelled & 50\% labelled \\
\midrule
Jing~\etal~\cite{jing2021videossl}$^{\ddagger}$ & 48.7 & 54.3 \\ 
Rizve~\etal~\cite{rizve2021defense_iclr} & 39.4 & 50.2 \\
\midrule
\textbf{MotionFit} (\textit{ours})      & 57.7 & 59.0 \\
\bottomrule
\end{tabular}
\footnotesize{$^{\ddagger}$ Use extra labels to pre-train a 2D CNN}.   \end{threeparttable}
}
\caption{\textbf{Comparison with semi-supervision} on video recognition at smaller scale. We report top-1 accuracy of models fine-tuned on 20\% (or 50\%) of UCF101 training data. Here $D_{\textit{MPLG}}$ (same as $D_{\textit{target}}$) is fraction of UCF101 train set and $D_{\textit{source}}$ is the UCF101 train set.}
\label{table:experiments:semis}
\end{table}

\section{Conclusion}
We propose a motion-augmented video self-training regime that transfers knowledge from motion to an appearance network. The trained model is motion-augmented and does not require costly optical flow computation during inference. Making it well suited for deployment on small-scale video datasets and video applications with constrained compute budgets. In order to improve the quality of the pseudo-labeling we introduce a simple yet effective multi-clip loss for training our pseudo-label generator. Our MotionFit provides a self-trained model that can be effectively fine-tuned on small-scale datasets for downstream tasks such as action recognition and clip retrieval. We fine-tune our model on two small-scale datasets and compare with state-of-the-art knowledge transfer, self-supervised and semi-supervised learning approaches that use the same amount of human labels for training. In all cases, our approach compares favorably to existing vision-only alternatives.

{\small
\bibliographystyle{ieee_fullname}
\bibliography{egbib}
}

\clearpage

\begin{center}
\textbf{\large Supplementary material for: \\ Motion-Augmented Self-Training for Video Recognition at Smaller Scale}
\end{center}

\setcounter{equation}{0}
\setcounter{figure}{0}
\setcounter{table}{0}
\setcounter{section}{0}
\makeatletter
\renewcommand{\theequation}{S\arabic{equation}}
\renewcommand{\thetable}{S\arabic{table}}
\renewcommand{\thefigure}{S\arabic{figure}}
\renewcommand{\thesection}{S\arabic{section}}

In this supplementary material, we first report qualitative results for video clip retrieval. Then we provide more analysis on semi-supervised action recognition. Finally, we make a larger comparison with self-supervised methods that use other than Kinetics dataset for training. 

\section{Qualitative results for video retrieval}
\begin{figure*}[t!]
    \centering
     \includegraphics[width=0.99\linewidth]{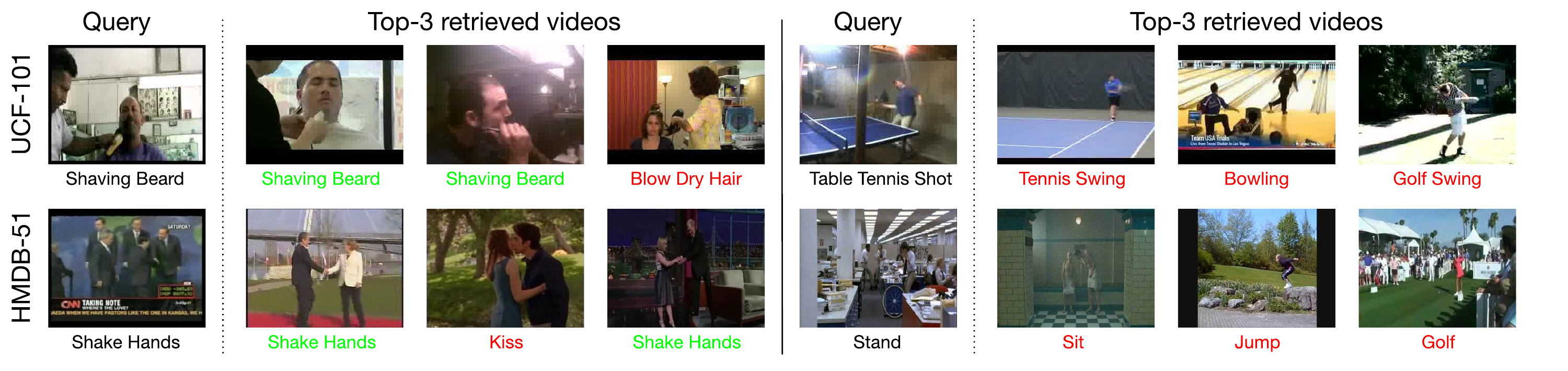}
    \caption{\textbf{Success and failure examples} for video retrieval. The left examples show top-retrieved videos belonging to the same action class as the query video. On the right, all three retrieved videos are from a different class. While our model may retrieve videos from different action classes, it still captures distinctive motion patterns like hand motion and human poses.} \label{supp:fig:failure_cases}
 \end{figure*}

To further investigate the quality of the learned representation, we illustrate a few success and failure examples in Figure~\ref{supp:fig:failure_cases}. Despite that some of the retrieved videos are from different action classes than the query video, the learned representation successfully captures similar motion patterns and not the appearance context. For example, on the first row the model captures hand motion, on the second row the model captures the dominant human poses.

\section{Semi-supervised action recognition}

\begin{table}[h!]
\centering
\resizebox{0.99\linewidth}{!}{
\begin{threeparttable}
\begin{tabular}{lccc}
\toprule
& \multicolumn{3}{c}{\textbf{UCF101, split 1}} \\
\cmidrule(lr){2-4}
 & 20\% labelled & 50\% labelled & $D_{\textit{source}}$ / Training dataset  \\
\midrule
Jing~\etal~\cite{jing2021videossl}$^{\ddagger}$ & 48.7 & 54.3 & training set\\ 
Rizve~\etal~\cite{rizve2021defense_iclr} & 39.4 & 50.2      & training set\\
\midrule
\textbf{MotionFit} (\textit{ours})      & 57.7 & 59.0       & training set \\
\textbf{MotionFit} (\textit{ours})      & 57.6 & 56.6       & training set $- D_{\textit{MPLG}}$ \\
\bottomrule
\end{tabular}
\footnotesize{$^{\ddagger}$ Use extra labels to pre-train a 2D CNN}.   \end{threeparttable}
}
\caption{\textbf{Comparison with semi-supervision} on video recognition at smaller scale. We report top-1 accuracy of models fine-tuned on 20\% (or 50\%) of UCF101 training data, which is also our $D_{\textit{MPLG}}$ (same as $D_{\textit{target}}$).}   
\label{supp:table:semis}
\end{table}

We further analyze our method for semi-supervised video recognition with ablation on the size of source dataset ($D_{\textit{source}}$) for self-training. The results are shown in Table~\ref{supp:table:semis}.
In the main paper, following competing methods, we use a labelled set (20\% or 50\% of training set) as $D_{\textit{MPLG}}$ and the full training set as $D_{\textit{source}}$. Here, we experiment by removing labelled samples from the training set to have a trimmed 
$D_{\textit{source}}=$ training set$- D_{\textit{MPLG}}$, with the remaining 80\% or 50\% samples of the training set. With this, the performance drops for both labeled subsets but the larger drop of 2.4\% for a 50\% labelled subset is due to the drastic decrease in the size of $D_{\textit{source}}$. While the 20\% labelled subset loses only 0.1\% as the data for self-training is only slightly reduced.
This further shows the importance of our self-training. 

\section{Comparison with self-supervised methods for action recognition}
In Table~\ref{supp:table:experiments:action_state_of_the_art} we list more results of self-supervised methods for action recognition that also utilize datasets other than Kinetics for training. Our method still outperforms other visual-only methods, including those that use a larger dataset for pre-training~\cite{DibaICCV2019}. 

\section{Comparison with self-supervised methods for clip retrieval}
In Table~\ref{supp:table:experiments:retrieval_state_of_the_art} we list more results of self-supervised methods for clip retrieval that also utilize other than Kinetics dataset for training. Our method still outperforms other visual-only methods.

\begin{table*}[t!]
\centering
\resizebox{0.78\linewidth}{!}{
\begin{tabular}{lccccc|cc}
\toprule
& \bfseries Dataset & \bfseries Backbone & \bfseries Frames & \bfseries Resolution & \bfseries Modality & \bfseries UCF101 & \bfseries HMDB51\\
\bottomrule
Sun~\etal~\cite{SunArxiv2019} & K400 & S3D & 16 & 112 & V + T & 79.5 & 44.6 \\
Owens~\etal~\cite{OwensECCV2018} & K400 & R3D-18 & 64 & 224 & V + A & 82.1 & - \\
Asano~\etal~\cite{AsanoArxiv2020} & K400 & R(2+1)D-18 & 30 & 112 & V + A & 83.1 & 47.1 \\
Asano~\etal~\cite{AsanoArxiv2020} & VGG-Sound~\cite{ChenICASSP2020} & R(2+1)D-18 & 30 & 112 & V + A & 87.7 & 53.1 \\
Korbar~\etal~\cite{KorbarNIPS2018} & K400 & MC3-18 & 25 & 224 & V + A & 85.8 & 56.9 \\
Korbar~\etal~\cite{KorbarNIPS2018} & Audioset~\cite{GemmekeICASSP2017} & MC3-18 & 25 & 224 & V + A & 89.0 & 61.6 \\
Alwassel~\etal~\cite{AlwasselArxiv2019} & K400 & R(2+1)D-18 & 32 & 224 & V + A & 86.8 & 52.6 \\
Alwassel~\etal~\cite{AlwasselArxiv2019} & Audioset~\cite{GemmekeICASSP2017} & R(2+1)D-18 & 32 & 224 & V + A & 93.0 & 63.7 \\
Alwassel~\etal~\cite{AlwasselArxiv2019} & IG-Kinetics~\cite{GhadiyaramCVPR2019} & R(2+1)D-18 & 32 & 224 & V + A & 95.5 & 68.9 \\
Xiao~\etal~\cite{XiaoArxiv2020} & K400 & SlowFast & 64 & 224 & V + A & 87.0 & 54.6 \\
Morgado~\etal~\cite{MorgadoArXiv2020} & K400 & R(2+1)D-18 & 32 & 224 & V + A & 87.5 & 60.8 \\
Morgado~\etal~\cite{MorgadoArXiv2020} & Audioset~\cite{GemmekeICASSP2017} & R(2+1)D-18 & 32 & 224 & V + A & 91.5 & 64.7 \\
Patrick~\etal~\cite{PatrickArxiv2020} & K400 & R(2+1)D-18 & 32 & 224 & V + A & 89.3 & 60.0 \\
Patrick~\etal~\cite{PatrickArxiv2020} & VGG-Sound~\cite{ChenICASSP2020} & R(2+1)D-18 & 32 & 224 & V + A & 89.4 & 62.1 \\
Patrick~\etal~\cite{PatrickArxiv2020} & Audioset~\cite{GemmekeICASSP2017} & R(2+1)D-18 & 32 & 224 & V + A & 92.5 & 66.1 \\
Patrick~\etal~\cite{PatrickArxiv2020} & IG-Kinetics~\cite{GhadiyaramCVPR2019} & R(2+1)D-18 & 32 & 224 & V + A & 95.2 & 72.8 \\
Miech~\etal~\cite{MiechCVPR2020} & HTM~\cite{MiechICCV2019} & S3D & 32 & 224 & V + T & 91.3 & 61.0 \\
Piergiovanni~\etal~\cite{PiergiovanniCVPR2020} & Youtube8M~\cite{AbuElHaijaArxiv2016} & S3D & 32 & 224 & V + T & 93.8 & 67.4 \\
\cmidrule{1-8}
ElNouby~\etal~\cite{ElNoubyArXiv2019} & UCF101 & R3D-18 & 16 & 112 & V & 64.4 & - \\
Kim~\etal~\cite{KimAAAI2019} & K400 & R3D-18 & 16 & 112 & V & 65.8 & 33.7 \\
Kong~\etal~\cite{KongNIPS2020} & K400 & R3D-18 & 8 & 112 & V & 69.4 & 37.8 \\
Luo~\etal~\cite{LuoVCP2020} & UCF101 & R(2+1)D-18 & 16 & 112 & V & 66.3 & 32.2 \\
Yao~\etal~\cite{YaoCVPR2020} & UCF101 & R(2+1)D-18 & 16 & 112 & V & 72.1 & 35.0 \\
Xu~\etal~\cite{XuCVPR2019} & UCF101 & R(2+1)D-18 & 16 & 112 & V & 72.4 & 30.9 \\
Cho~\etal~\cite{ChoArxiv2020} & UCF101 & R(2+1)D-18 & 16 & 112 & V & 74.8 & 36.8 \\
Han~\etal~\cite{HanICCVW2019} & K400 & R-2D3D-34 & 25 & 224 & V & 75.7 & 35.7 \\
Jing~\etal~\cite{JingArxiv2018} & K400 & R3D-18 & 64 & 112 & V & 76.6 & 47.0 \\
Zhuang~\etal~\cite{ZhuangCVPR2020} & K400 & SlowFast & 16 & 112 & V & 77.0 & 46.5 \\
Han~\etal~\cite{HanArxiv2020} & K400 & R-2D3D-18 & 25 & 224 & V & 78.1 & 41.2 \\
Benaim~\etal~\cite{BenaimCVPR2020} & K400 & S3D-G & 64 & 224 & V & 81.1 & 48.8 \\
Han~\etal~\cite{HanNIPS2020} & UCF101 & S3D & 32 & 128 & V & 81.4 &  52.1 \\
Han~\etal~\cite{HanNIPS2020} & K400 & S3D & 32 & 128 & V & 87.9 &  54.6 \\
Diba~\etal~\cite{DibaICCV2019} & Youtube8M~\cite{AbuElHaijaArxiv2016} & STCNet & 32 & 112 & V & 88.1 & 59.9 \\
\cmidrule{1-8}
\textbf{MotionFit} (\textit{ours})  & K400 & R(2+1)D-18 & 32 & 112 & V & 88.9 & 61.4 \\
\textbf{MotionFit} (\textit{ours})  & K400 & S3D-G & 64 & 224 & V & 90.1 & 50.6 \\
\bottomrule
\end{tabular}
}
\caption{\textbf{Comparison with self-supervised methods on video action recognition.} We report top-1 accuracy of fine-tuned models averaged over all 3 splits of UCF101 and HMDB51. Our approach is the best when only considering the visual modality (V).  
}
\label{supp:table:experiments:action_state_of_the_art}
\end{table*}

\begin{table*}[t!]
\centering
\resizebox{0.75\linewidth}{!}{
\begin{tabular}{lcccccccccc}
\toprule
& & & & \multicolumn{3}{p{3.0cm}}{\centering\bfseries UCF101} & \multicolumn{3}{p{3.0cm}}{\centering\bfseries HMDB51} \\
\cmidrule(lr){5-7} \cmidrule(lr){8-10}
& {\bfseries Dataset} & {\bfseries Backbone} & {\bfseries Modality} & R@1 & R@5 & R@20 & R@1 & R@5 & R@20 \\
\midrule
Asano~\etal~\cite{AsanoArxiv2020} & K400 & R(2+1)D-18 & V + A & 52.0 & 68.6 & 84.5 & 24.8 & 47.6 & 75.5 \\
Patrick~\etal~\cite{PatrickArxiv2020} & K400 & R(2+1)D-18 & V + A & 57.4 & 73.4 & 88.1 & 25.4 & 51.4 & 75.0 \\
\midrule
Xu~\etal~\cite{XuCVPR2019} & UCF101 & R(2+1)D-18 & V & 10.7 & 25.9 & 47.3 & 5.7 & 19.5 & 45.8 \\
Benaim~\etal~\cite{BenaimCVPR2020} & K400 & S3D-G & V & 13.0 & 28.1 & 49.5 & - & - & - \\
Noroozi~\etal~\cite{NorooziECCV2016} & UCF101 & AlexNet & V & 19.7 & 28.5 & 40.0 & - & - & - \\
Luo~\etal~\cite{LuoVCP2020} & UCF101 & R(2+1)D-18 & V & 19.9 & 33.7 & 50.5 & 6.7 & 21.3 & 49.2 \\
Han~\etal~\cite{HanArxiv2020} & UCF101 & R(2+1)D-18 & V & 20.2 & 40.4 & 64.7 & 7.7 & 25.7 & 57.7 \\
Yao~\etal~\cite{YaoCVPR2020} & UCF101 & R(2+1)D-18 & V & 20.3 & 34.0 & 51.7 & 8.2 & 25.3 & 51.0 \\
Kong~\etal~\cite{KongNIPS2020} & K400 & R3D-18 & V & 22.0 & 39.1 & 56.3 & - & - & - \\
Cho~\etal~\cite{ChoArxiv2020} & UCF101 & R3D-18 & V & 24.6 & 41.9 & 62.7 & 10.3 & 26.6 & 54.6 \\
Buchler~\etal~\cite{BchlerECCV2018} & UCF101 & CaffeNet & V & 25.7 & 36.2 & 49.2 & - & - & - \\
Han~\etal~\cite{HanNIPS2020} & UCF101 & S3D & V & 53.3 & 69.4 & 82.0 & 23.2 & 43.2 & 65.5 \\
\midrule
\textbf{MotionFit} (\textit{ours}) & K400 & S3D-G & V & 31.6 & 51.7 & 70.3 & - & - & - \\
\textbf{MotionFit} (\textit{ours}) & K400 & R(2+1)D-18 & V & 61.6 & 75.6 & 85.5 & 29.4 & 46.5 & 66.7 \\
\bottomrule
\end{tabular}
}
\caption{\textbf{Comparison with self-supervised methods on video clip retrieval.} We report recall values R@$n$ for $n$ = 1, 5, 20 on UCF101 and HMDB51 split 1. Our approach is best when only considering the visual modality and on par with methods that use an additional audio modality during training. }
\label{supp:table:experiments:retrieval_state_of_the_art}
\end{table*}

\end{document}